\documentclass{article}

\usepackage[preprint]{neurips_2020}

\usepackage[utf8]{inputenc} 
\usepackage[T1]{fontenc}    
\usepackage{hyperref}       
\usepackage{url}            
\usepackage{booktabs}       
\usepackage{amsfonts}       
\usepackage{nicefrac}       
\usepackage{microtype}      
\usepackage{chessboard}
\usepackage{float}
\usepackage{pgfplots}
\usepackage{caption}
\usepackage{amsmath}
\usepackage{subcaption}
\usepackage{filecontents}
\usepackage{siunitx}
\usepackage{natbib}

\title{Playing Chess with Limited Look Ahead}

\author{%
  Arman Maesumi \\
  Department of Computer Science\\
  University of Texas at Austin\\
  \texttt{arman@cs.utexas.edu} \\
}

\begin{document}

\maketitle

\begin{abstract}
We have seen numerous machine learning methods tackle the game of chess over the years. However, one common element in these works is the necessity of a finely optimized look ahead algorithm. The particular interest of this research lies with creating a chess engine that is highly capable, but restricted in its look ahead depth. We train a deep neural network to serve as a static evaluation function, which is accompanied by a relatively simple look ahead algorithm. We show that our static evaluation function has encoded some semblance of look ahead knowledge, and is comparable to classical evaluation functions. The strength of our chess engine is assessed by comparing its proposed moves against those proposed by Stockfish. We show that, despite strict restrictions on look ahead depth, our engine recommends moves of equal strength in roughly $83\%$ of our sample positions.
\end{abstract}

\section{Problem statement and motivation}
The application of machine learning in game playing has been a popular testing bed for the pattern recognition capabilities of artificial learning systems. Over the years we have seen machine learning systems tackle the game of chess, with the recent contributions being Giraffe \citep{giraffe}, DeepChess \citep{deepchess}, and AlphaZero \citep{alphazero}. These chess engines use a variety of learning-based methods to play the game, but one common element is the necessity of a finely optimized look ahead algorithm. For instance, Giraffe and DeepChess utilize a probabilistic variant of alpha-beta pruning, and AlphaZero uses Monte Carlo Tree Search (MCTS). The particular interest of this research lies with creating a chess engine that is highly capable, but restricted in its look ahead depth.  

Human chess grandmasters excel at accurately assessing the strength of a particular position. Strong human players distinguish themselves from chess engines through their \textit{static evaluation} ability. Evaluating a position \textit{statically} means that you do not consider future moves in your evaluation. Classical chess engines use handcrafted static evaluation functions that are optimized by chess experts. We would like to train an artificial neural network that acts as a highly capable static evaluation function. We hope to utilize the capability of neural networks to encode look ahead information inside of the static evaluation function, thus making its predictions more potent than classical evaluation functions.

\section{Prior work}

While computer chess has been explored extensively, there are very few papers that specifically focus on playing the game with limited look ahead. The most notable work that addresses this is by \citet{sabatelli}, which is the primary influence of this research. In this work, a supervised learning approach was used to train an artificial neural network that served as a chess board evaluator. The network was trained on board configurations that were labelled by a classical chess engine. They also explored various board representations and neural architectures, which we refer to later. The key differences between this research and Sabatelli's paper can be seen in the dataset, and our use of a deep autoencoder. We also employ a look ahead algorithm that uses alpha-beta pruning. Sabatelli's model was evaluated by playing $30$ games against humans on an online chess website. The paper claims that the playing strength of the model is roughly $2000$ on the Elo scale; although, the sample size is quite low.

\section{Classical evaluation methods}
Before machine learning methods were a viable option, the primary technique used in computer chess was minimax coupled with a handcrafted heuristic function. Minimax is a rudimentary depth-first look ahead algorithm that evaluates every future position to a specified depth. As we know, brute-force methods do not work in chess, as the search space is far too large. This gave rise to alpha-beta pruning, which is a variant of minimax that prunes out much of the search space. For the sake of brevity we will summarize the techniques used in \textit{Stockfish} \citep{stockfish}, the strongest classical chess engine. The static evaluation function used by Stockfish is a linear combination of several factors \citep[see][]{evaluation}:

\begin{itemize}
    \item Pawn Structure: Specific arrangements of pawns can greatly impact the evaluation of a particular position. For example, a passed pawn (one which will likely obtain promotion), is very valuable, where as a backwards pawn (one which cannot advance safely), is less valuable.
    \item Material: Each individual piece is given a value that describes their strength relative to a single pawn. Typically, a queen, rook, bishop, and knight are worth $9$, $5$, $3$, and $3$ pawns respectively. However, strong engines like Stockfish may dynamically alter these piece values depending on the position at hand.
    \item King Safety: The ultimate goal in chess is to successfully attack the opponent's king. Therefore, the safety of one's king is of utmost importance. Stockfish will penalize an exposed king in the evaluation, but reward a king that is well protected.
    \item Mobility: It is often the case that having more squares at your disposal is beneficial. For instance, a bishop with only $2$ available moves is typically worth much less than a bishop that slices the entire board along a diagonal. The mobility score is calculated as a function of the number of available moves that each piece has. This is often done on a piece-by-piece basis, as some piece types have different mobility patterns.
    \item Piece-Square Tables: Each piece type is awarded a bonus (or penalty) depending on their location. For example, in the early game it is beneficial to push pawns to the center of the board to control more space. The piece-square tables encourage this type of piece development.
    \item Evaluation of Pieces: Similar to the piece-square tables, each piece is awarded a bonus or penalty depending on a few specific criteria. For instance, rooks enjoy being on open files, and knights greatly benefit from being placed in pawn outposts.
    \item Tempo: It is typically the case that accomplishing a goal in the least number of moves is beneficial. Playing moves that do not have any real effect on the game will most likely result in loss of initiative. Stockfish will penalize moves that cause a loss of tempo/initiative.
    \item Center Control: Occupying squares near the center of the board is a general strategy in chess theory. Controlling more space in the center allows the player to both defend their territory, and attack the opponent's territory easier.
\end{itemize}{}

Most classical evaluation functions use \textit{centipawns} as their unit of measure. A centipawn represents $1/100$th of a pawn. For instance, an evaluation of $+325cp$ indicates that white has an effective advantage of $3.25$ pawns. A negative evaluation indicates an advantage for black. It is important to note that the centipawn score assigned to a particular position is dependant on search depth. However, the static evaluation function serves to approximate the centipawn score by using no search at all.

\begin{figure}[H]
    \centering

    \chessboard[setfen=5rk1/1pp2bpp/p1np4/P3p3/1P2P2P/1R2P3/2P3q1/2KQ1R2 b - - 0 1, showmover, pgfstyle=straightmove, linewidth=0.075em, markmoves={f7-b3}, boardfontsize=14pt,labelfontsize=8pt]

    \caption{At depth $23$, Stockfish says this position favors black with an evaluation of $-990cp$. Black is preparing to capture white's rook on $b3$, resulting in a helpless position for white.}
    \label{fig:centipawn_example}
\end{figure}{}

\section{The dataset}

Similar to the dataset proposed by \citet{sabatelli}, we extract unique board configurations from public chess databases. The board configurations are labelled using Stockfish 11, where each label represents the centipawn advantage relative to white. The dataset in this research is unique in two ways. First, we created a dataset that is considerably larger than what was used in the prior research. Instead of training on $3$ million samples, we train on upwards of $20$ million samples. Secondly, our dataset features what we call ``artificial dataset expansion.'' The idea is simple, we take a portion of our sample board configurations, make a random legal move, then evaluate the resulting board using Stockfish. A similar idea was proposed by \citet{giraffe}, where it was used to provide the neural network with a better understanding of various positions that humans would generally ignore. Therefore, artificial dataset expansion is not used due to a lack of data, but rather to increase the general understanding of positional chess in our models. Prior to expanding our dataset, the distribution of centipawn scores resembles a very narrow gaussian. This is because chess positions remain quite drawish until one player makes a mistake. By applying random moves to the boards in our dataset, we are able to force the drawish games to have higher evaluation. The expanded dataset will reduce bias in our models by making the distribution of centipawn scores less narrow. Below is a histogram that shows the distribution of centipawn scores. We can see that the large majority of scores are in the range $-250 \leq cp \leq 250$. The noticeable peaks at $\pm10000cp$ are caused by the large number of positions where checkmate is identified by the engine.

\begin{figure}[H]
    \centering
   \begin{tikzpicture}
    \begin{axis}[
        ymin=0, ymax=8000000,
        scaled x ticks=false,
        minor y tick num = 3,
        area style,
        ]
    \addplot+[ybar interval,mark=no] plot coordinates { 
    ( -10001.0 , 1184256.0 )
    ( -9600.96 , 48.0 )
    ( -9200.92 , 152.0 )
    ( -8800.88 , 560.0 )
    ( -8400.84 , 1352.0 )
    ( -8000.8 , 6608.0 )
    ( -7600.76 , 22176.0 )
    ( -7200.719999999999 , 20864.0 )
    ( -6800.68 , 46176.0 )
    ( -6400.639999999999 , 64352.0 )
    ( -6000.6 , 33448.0 )
    ( -5600.5599999999995 , 19032.0 )
    ( -5200.5199999999995 , 3056.0 )
    ( -4800.48 , 1512.0 )
    ( -4400.44 , 1208.0 )
    ( -4000.3999999999996 , 3912.0 )
    ( -3600.3599999999997 , 10640.0 )
    ( -3200.3199999999997 , 28192.0 )
    ( -2800.2799999999997 , 70744.0 )
    ( -2400.24 , 152648.0 )
    ( -2000.1999999999998 , 367336.0 )
    ( -1600.1599999999999 , 1018080.0 )
    ( -1200.119999999999 , 1544072.0 )
    ( -800.0799999999999 , 3318504.0 )
    ( -400.03999999999905 , 6428800.0 )
    ( 0.0 , 6514424.0 )
    ( 400.0400000000009 , 3241992.0 )
    ( 800.0799999999999 , 1555008.0 )
    ( 1200.1200000000008 , 1015352.0 )
    ( 1600.1599999999999 , 368896.0 )
    ( 2000.2000000000007 , 154216.0 )
    ( 2400.24 , 70832.0 )
    ( 2800.2800000000007 , 27600.0 )
    ( 3200.3200000000015 , 10664.0 )
    ( 3600.3600000000006 , 3760.0 )
    ( 4000.4000000000015 , 1288.0 )
    ( 4400.4400000000005 , 1424.0 )
    ( 4800.480000000001 , 2864.0 )
    ( 5200.52 , 17952.0 )
    ( 5600.560000000001 , 32096.0 )
    ( 6000.6 , 62792.0 )
    ( 6400.639999999999 , 43640.0 )
    ( 6800.68 , 20000.0 )
    ( 7200.720000000001 , 21168.0 )
    ( 7600.760000000002 , 6632.0 )
    ( 8000.799999999999 , 1448.0 )
    ( 8400.84 , 640.0 )
    ( 8800.880000000001 , 104.0 )
    ( 9200.920000000002 , 48.0 )
    ( 9600.960000000003 , 1224280.0 )
    ( 10001.0 , 1283152.0 )
    };
    \end{axis}
    \end{tikzpicture}
    \label{fig:centipawn_dist_after}
\caption{The centipawn score distribution after using artificial dataset expansion.}
\end{figure}

\section{Board representation}

Representing a chess board numerically comes with a few notable challenges. We must represent $64$ tiles in which there are $12$ unique piece types present:  king, queen, rook, bishop, knight, and pawn, each of which has a white and black variant. Thus, we have $12\times64 = 768$ categorical variables for the pieces. However, we would like to provide more information to the model, so we also include the following features:

\begin{itemize}
    \item Turn bit: A single bit that indicates who's turn it is.
    \item Castling bits: Four bits that represent the castling rights of white and black.
    \item Check bits: Two bits that represent if white or black are in check.
\end{itemize}{}

In total, our board representation is a flat vector with $775$ binary features. We use the \textit{bitboard} representation to denote the locations of each piece. This means the first 64 features correspond to the locations of white pawns, the next 64 features correspond to white knights, and so on. For instance, the bitboard for the following arrangement of pawns would be:

\begin{figure}[h]
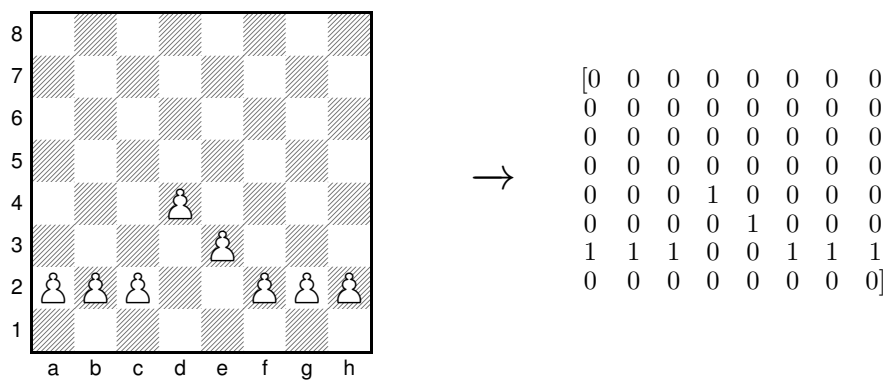

\centering
\begin{subfigure}{.5\textwidth}
    \centering

    \chessboard[setfen=8/8/8/8/3P4/4P3/PPP2PPP/8 w - - 0 1, showmover=false, boardfontsize=16pt,labelfontsize=8pt]
\end{subfigure}
{\LARGE$\xrightarrow{}$}
\begin{subfigure}{.40 \textwidth}
    \centering
    $\begin{matrix}{}
    [0 & 0 & 0 & 0 & 0 & 0 & 0 & 0\\
    0 & 0 & 0 & 0 & 0 & 0 & 0 & 0\\
    0 & 0 & 0 & 0 & 0 & 0 & 0 & 0\\
    0 & 0 & 0 & 0 & 0 & 0 & 0 & 0\\
    0 & 0 & 0 & 1 & 0 & 0 & 0 & 0\\
    0 & 0 & 0 & 0 & 1 & 0 & 0 & 0\\
    1 & 1 & 1 & 0 & 0 & 1 & 1 & 1\\
    0 & 0 & 0 & 0 & 0 & 0 & 0 & 0]\\
    \end{matrix}$
\end{subfigure}
\caption{The bitboard representation of white pawns. Note that the matrix on the right is one dimensional.}
\end{figure}

Each piece type is assigned its own bitboard. These $12$ bitboards are then concatenated to form the board representation. The turn, castling, and check bits are concatenated as well.

It is worth noting that there are many other ways to represent a chess board numerically. For instance, \citet{giraffe} proposed a novel coordinate representation that did not rely on bitboards, but rather assigned $x,y$ coordinates to each present piece. One advantage of this representation is that the feature vector is noticeably smaller (roughly half the size) than the bitboard representation. However, we found that the bitboard representation yielded stronger playing strength.

\section{Dimensionality reduction}

The feature vector produced by our bitboard representation is quite sparse. This is because there are $64$ categorical features, each of which encodes $12$ features. On average, most tiles on a chess board will be empty, as there are at most $32$ pieces on the board. Naturally, we would like to compress this sparse feature vector into a significantly smaller latent space. To do this, we employ a deep autoencoder, similar to the deep belief network used by \citet{deepchess}. The autoencoder consists of seven fully connected layers with tanh activations. The output layer uses a sigmoid activation. The following hyperparameters were used:

\begin{itemize}
    \item Number of units: ($775\rightarrow512\rightarrow256\rightarrow128\rightarrow256\rightarrow512\rightarrow775)$.
    \item Dropout \citep[see][]{dropout}: A probability of $25\%$ was used between all hidden layers.
    \item Optimizer: Adam optimizer was used with $\eta = 0.001$, $\beta_1 = 0.90$, $\beta_2 = 0.999$ \citep{adam}.
    \item Loss Function: Binary crossentropy was used,
    $$L(y, \hat{y}) = -\frac{1}{N}\sum_{i=0}^N(y\log \hat{y_i} + (1-y)\log(1-\hat{y_i}))$$
\end{itemize}{}
The autoencoder was trained on $22$ million board configurations, and tested on $3$ million. These board configurations were chosen uniformly at random from the total dataset. Information regarding source code and training hardware can be found in Appendix C. The testing loss is shown below.

\begin{figure}[H]
    \centering
    \begin{tikzpicture}
    \begin{axis}[
        xlabel={Epoch},
        ylabel={Loss}
    ]
    \addplot table [x=epoch, y=loss, col sep=comma] {encoder_loss.csv};
    \end{axis}
    \end{tikzpicture}
    \caption{The decoded loss at every epoch on the testing dataset. Note that the scale is $10^{-3}$, so our decoded board representations are almost always perfectly accurate.}
    \label{fig:encoder_graph}
\end{figure}
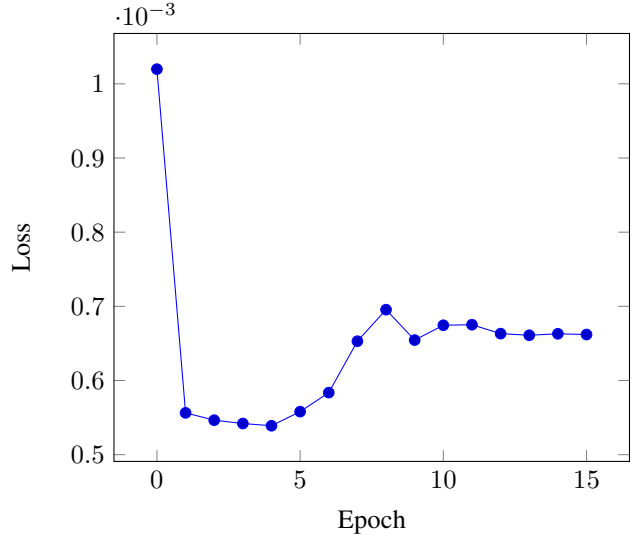{}

\section{The classifier}

Given a particular chess position, we would like to assess the probability that white is winning, black is winning, and that there is no significant advantage. Querying such a predictive model will allow our look ahead algorithm to easily prune trivially bad branches. Recall that the dataset we are working with is a collection of unique board configurations and their corresponding Stockfish evaluation (in centipawns). We can easily assign one-hot encoded categorical labels by using the following transformation

\[ \textrm{LABEL}(cp) = \begin{cases} 
      \textrm{Black winning} & cp \leq -150 \\
      \textrm{Drawish} & -150\leq cp \leq 150 \\
      \textrm{White winning} & 150\leq cp 
   \end{cases}
.\]

The choice of $\pm150cp$ as the decision boundary comes from chess theory. To perform this prediction, we employ a deep neural network with the following hyperparameters:

\begin{itemize}
    \item Number of units: ($775\rightarrow1024\rightarrow512\rightarrow256\rightarrow3)$.
    \item Dropout: A probability of $25\%$ was used between all hidden layers.
    \item Optimizer: Adam optimizer was used with $\eta = 0.001$, $\beta_1 = 0.90$, $\beta_2 = 0.999$.
    \item Loss Function: Categorical crossentropy was used, $$L(y,\hat{y}) = -\sum_{j=0}^M\sum_{i=0}^N(y_{ij} \log\hat{y_{ij}}).$$
\end{itemize}{}

ReLU activations are used between all hidden layers, and a softmax activation is used for the output. This model was trained on 20 million boards, and tested on 2 million. The testing accuracy is shown below. 

\begin{figure}[H]
    \centering
    \begin{tikzpicture}
    \begin{axis}[
        xlabel={Epoch},
        ylabel={Testing Accuracy}
    ]
    \addplot table [x=epoch, y=acc, col sep=comma] {class_acc.csv};
    \end{axis}
    \end{tikzpicture}
    \caption{The classification accuracy at every epoch on the testing dataset.}
    \label{fig:classification_graph}
\end{figure}
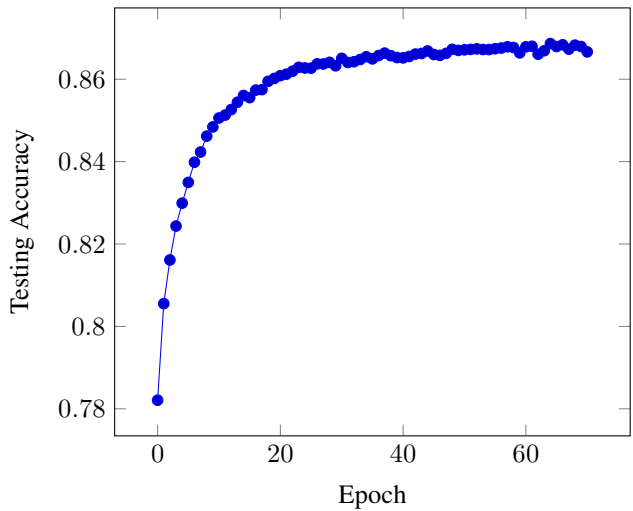{}

It is worth noting that many of our testing samples are relatively close to the decision boundary of $\pm150cp$. To fully realize the effect that these samples have on our accuracy, we tested the model on the same dataset, but removed boards that have evaluations in the range $115 \leq cp \leq 185$ or $-185 \leq cp \leq -115$. The accuracy on this modified testing set is considerably better, and achieves upwards of $93\%$ classification accuracy. To take it one step further, we did another test where all boards with evaluations in the range $-250 \leq cp \leq 250$ were removed (all ``drawish" boards are gone at this point). The classification accuracy under these restrictions increases to $96\%$. This means our model is highly capable at assessing boards that contain a non-negligible advantage, which is exactly what is needed for pruning.

\section{The static evaluation function}

The core component of any chess engine is, of course, the evaluation function. This model will be quite similar to the classifier; however, we will be training directly on the centipawn evaluations given by Stockfish. Before training our deep neural network, we must normalize the centipawn scores in our dataset. To do this, we first place a limit on the magnitude of our centipawn scores. Evaluations are capped between $5000cp$ and $-5000cp$, which is equivalent to $\pm 50$ pawns. Conceptualizing what a ``$50$ pawn advantage'' actually means is quite difficult because at this point the magnitude of the advantage is blurred and ultimately not important. Since our model uses a tanh activation function in its output unit, we must normalize our evaluations to the range $[-1,1]$. To do this, we simply apply the following transformation to each evaluation, $\mathcal{E}$, in our dataset

$$ \textsc{NORMALIZE}(\mathcal{E}) = 2* \frac{\mathcal{E} - \textrm{min}}{\textrm{max} - \textrm{min}} - 1, \textrm{where max} = 5000 \hspace{0.5em} \textrm{and} \hspace{0.5em} \textrm{min} = -5000.$$

With the normalized evaluations, we may now train the deep neural network. This is inherently a more difficult task, so we will be using a bigger model with the following parameters.

\begin{itemize}
    \item Number of units: ($128\rightarrow2048\rightarrow2048\rightarrow2048\rightarrow1)$.
    \item Dropout: A probability of $30\%$ was used between all hidden layers.
    \item Optimizer: Adam optimizer was used with $\eta = 0.001$, $\beta_1 = 0.90$, $\beta_2 = 0.999$.
    \item Loss Function: Mean squared error (MSE) was used, $$L(y,\hat{y})=\frac{1}{N}\sum_{i=1}^N(y_i - \hat{y}_i)^2.$$
\end{itemize}{}

ReLU activations are used between all hidden layers, and a tanh activation is used for the output. Note that we are now using the output of the deep autoencoder as the input to this model. This is why the input layer has $128$ units, rather than $775$. This model was also trained on 20 million boards, and tested on 2 million. The testing loss is shown below. 

\begin{figure}[H]
    \centering
    \begin{tikzpicture}
    \begin{axis}[
        xlabel={Epoch},
        ylabel={Testing Loss}
    ]
    \addplot table [x=epoch, y=loss, col sep=comma] {reg_loss.csv};
    \end{axis}
    \end{tikzpicture}
    \caption{The loss at every epoch on the testing dataset. Note that the scale of the graph is $10^{-2}$.}
    \label{fig:regression_graph}
\end{figure}
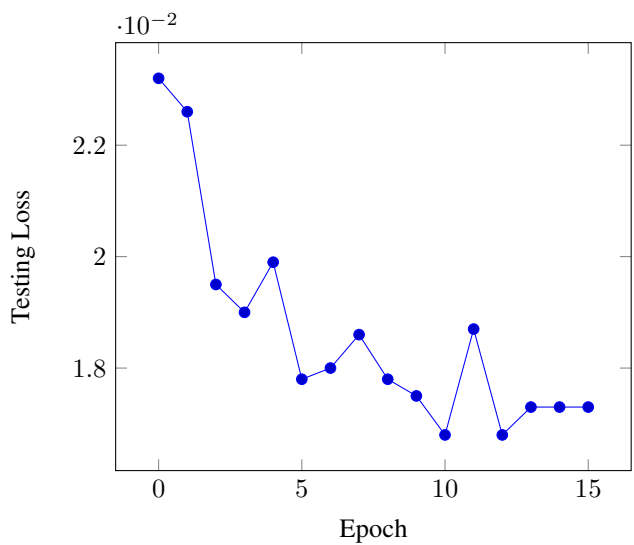{}

We can see that the mean squared error converges to roughly $0.017$. To properly put this number into perspective, we can reverse the normalization to arrive at $85cp$. This means on average the model's prediction is off by $0.85$ pawns. Note that the labels in our dataset come from Stockfish at depth $12$, so we can conclude that the model has encoded some look ahead information in its evaluation, which is precisely the goal of this research.

We have provided a few sample outputs from the trained static evaluation function (see Appendix A). There are a variety of positions, containing both small and large advantages. The output of our model and the output of Stockfish at depth $23$ are shown. We can see that our model accurately assess these positions without using any look ahead algorithm. This further reinforces the idea that our model has, to some degree, encoded look ahead information in its layers.

\section{Look ahead and playing strength}\label{lookahead}

While our static evaluation model is quite accurate in most situations, it is possible for the model to give a wildly false prediction. This is true for any static evaluation function, and is precisely what motivates the usage of a look ahead algorithm. We implemented a variant of alpha-beta pruning that utilizes our model as its core heuristic. The search algorithm was written in Golang for the sake of speed. It features multi-threaded search, move ordering, transposition tables, iterative deepening, and our models (which run on the GPU) as the search heuristic. Using an NVIDIA GTX 1080 Ti, and Intel Core i9-9900k, the search algorithm is able to process \num[group-separator={,}]{25000} to \num[group-separator={,}]{40000} positions per second. Using a neural network as a search heuristic is much more costly compared to a traditional evaluation function, but we make up for this through the potency of our model.

To test the strength of our engine, we randomly sampled $500$ board configurations from the testing set, and ran them through our search algorithm at depth $5$. To gauge the strength of our algorithm, the same samples were evaluated using Stockfish at depth $23$. We found that $83\%$ of the moves chosen by our engine were of equal strength to the moves chosen by Stockfish. These results are very promising, as the search depth used by our algorithm was considerably lower than that of Stockfish. A sample of the positions used in this experiment can be found in Appendix B.

\section*{Future research directions}
We would like to explore how differently our engine performs with the Monte Carlo Tree Search algorithm, rather than the alpha-beta pruning algorithm implemented in (\ref{lookahead}). Secondly, training a variational autoencoder (VAE) to generate possible ``future positions'' from a given position would be an interesting avenue to explore. Perhaps this can serve as a novel ``neural look ahead'' algorithm. Finally, we would like to apply network distillation to the static evaluation function, as this would greatly increase inference speed in the look ahead algorithm.

\section*{Broader Impact}
While the work presented in this paper will not effect the daily lives of most people, it can certainly effect those who play chess competitively. Due to computer dominance in chess, it is sadly very easy to cheat in online chess matches. Cheating in such events often involves consulting an engine, such as the one presented in this work. However, we hope that our work is used productively in a more instructive setting. For example, using this work to practice, or study the game of chess.

\begin{ack}
This research was conducted over a one-year span during which I benefited from the input of Dr. Bajaj, Dr. Plaxton, and Dr. Bulko all of whom are professors of computer science at the University of Texas at Austin. I am particularly thankful for Dr. Bajaj's suggestion to submit this research to NeurIPS.
\end{ack}

\bibliography{bibliography}
\section*{Appendix A}

\begin{figure}[H]
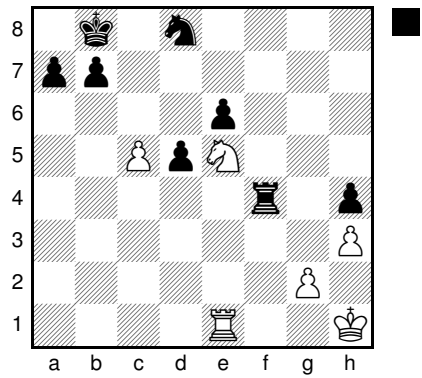
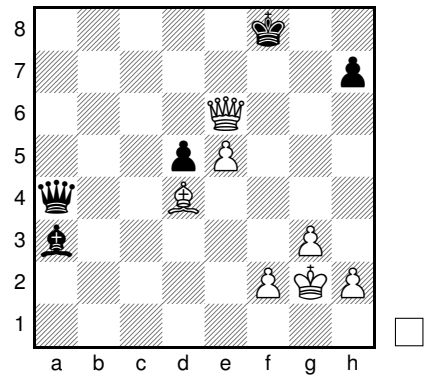
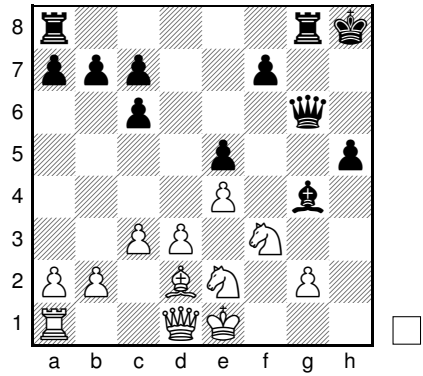
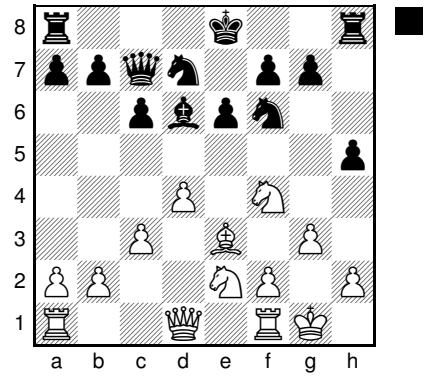
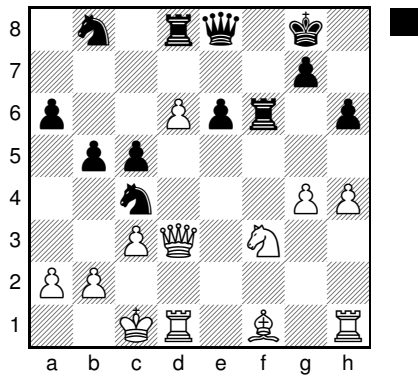

\centering
\begin{subfigure}{.5\textwidth}
    \centering
    \chessboard[setfen=1k1n4/pp6/4p3/2PpN3/5r1p/7P/6P1/4R2K b - - 4 46, showmover, boardfontsize=16pt,labelfontsize=8pt]
    \caption{Prediction: $-310cp$, Stockfish: $-420cp$}
\end{subfigure}%
\begin{subfigure}{.5\textwidth}
    \centering
    \chessboard[setfen=5k2/7p/4Q3/3pP3/q2B4/b5P1/5PKP/8 w - - 1 37, showmover, boardfontsize=16pt,labelfontsize=8pt]
    \caption{Prediction: $4400cp$, Stockfish: $5000cp$}
\end{subfigure}
\begin{subfigure}{.5\textwidth}
    \centering
    \chessboard[setfen=r5rk/ppp2p2/2p3q1/4p2p/4P1b1/2PP1N2/PP1BN1P1/R2QK3 w - - 1 21, showmover, boardfontsize=16pt,labelfontsize=8pt]
    \caption{Prediction: $440cp$, Stockfish: $400cp$}
\end{subfigure}%
\begin{subfigure}{.5\textwidth}
    \centering
    \chessboard[setfen=r3k2r/ppqn1pp1/2pbpn2/7p/3P1N2/2P1B1P1/PP2NP1P/R2Q1RK1 b kq - 1 15, showmover, boardfontsize=16pt,labelfontsize=8pt]
    \caption{Prediction: $-170cp$, Stockfish: $-180cp$}
\end{subfigure}
\begin{subfigure}{.5\textwidth}
    \centering
    \chessboard[setfen=1n1rq1k1/6p1/p2Ppr1p/1pp5/2n3PP/2PQ1N2/PP6/2KR1B1R b - - 1 26, showmover, boardfontsize=16pt,labelfontsize=8pt]
    \caption{Prediction: $-105cp$, Stockfish: $-110cp$}
\end{subfigure}

\caption[short]{A few samples comparing our static evaluation model, and Stockfish at depth $23$.}
\end{figure}
\newpage
\section*{Appendix B}
Below are a subsample of the $500$ board configurations used to compare our engine to Stockfish. The boards are given in Forsyth–Edwards Notation (FEN). The first listing contains boards where our engine's candidate move agreed with Stockfish. The second listing contains the boards that were not agreed upon.

Listing one:
\begin{itemize}
\item r2q1rk1/1pp1n1pp/1b2b3/pP1p4/P2Pp3/B1P1PB1P/6PN/R2Q1RK1 w - - 0 18
\item r5k1/1p4pp/2p1q3/pP1pbn2/P2Pp3/B1P1P2P/6P1/R3Q1K1 w - - 0 25
\item r5k1/1p4pp/1bp1q3/pP1p1n2/P2Pp3/B1P1P2P/6PN/R3Q1K1 w - - 0 23
\item 2rqk2r/pp3p1p/2nb1np1/1B1p1p2/8/1PN1PQ2/PB1P1PPP/R3K2R w KQk - 0 12
\item 2rqk2r/pp3p1p/2nb2p1/1B1n1p2/8/1P2PQ2/PB1P1PPP/R3K2R w KQk - 0 13
\item 2rr2k1/pp3p1p/6p1/1B2np2/1b5P/1P2P3/PB1P1PP1/3RK2R w K - 2 17
\item r2qr1k1/pp3ppp/6b1/2bNp1Pn/2Pn4/3PBN1P/PP3PB1/R2QR1K1 w - - 3 16
\item 4r1k1/pp1q1ppp/3b2b1/3N2Pn/1PPp4/3P1Q1P/P2B1PB1/4R1K1 w - - 1 21
\item 6k1/pp3pp1/6bp/6P1/1PPp1Q2/3P3P/P4PBK/4q3 w - - 0 26
\item 6k1/pp2qp2/7p/7P/1PP2Q2/3b1B2/P4PK1/8 w - - 0 34
\item r2q1rk1/1p2bppp/2p1p3/p1nnP3/8/P5P1/1PPNQPBP/R1B2RK1 w - - 0 14
\item 5bk1/1p1q1p1p/2p1p1pP/p1n1PnB1/2P5/P4NP1/1P2QPB1/6K1 w - - 1 28
\item Q4bk1/5p1p/2p1p1pP/1pB5/2P1q1P1/P7/1P3PK1/8 w - - 4 43
\item r2r4/2n1kppp/1ppNpn2/2P5/pP2P3/P5P1/5PBP/2RR1K2 w - - 1 25
\item r2r4/4k1pp/1nRNpp2/3nP3/pP6/P5P1/5PBP/2R2K2 w - - 0 29
\item r4rk1/1p1n2p1/1qp1p2p/p5P1/2pP2n1/4PNB1/PPQR1P2/2K4R w - - 0 19
\item rn2kb1r/pp2ppp1/1qp2n1p/3p4/3P2bB/3BP3/PPPN1PPP/R2QK1NR w KQkq - 3 7
\item r4rk1/1p2qppp/2n5/3np3/p3P3/5P2/PP1B2PP/2RQKR2 w - - 0 19
\item r5k1/1p2qppp/8/4p3/p2r4/5P2/PP3KPP/2RQ1R2 w - - 0 23
\item r2qk2r/4bp1p/p1bp1p2/1p2p3/4P2N/2NQ4/PPP2PPP/2KR3R w kq - 0 13
\item 2r1k2r/2q1bp1p/p2p1p2/3QpN2/1p2P3/8/PPP2PPP/1K1R3R w k - 3 17
\item 2r1k3/5p1p/p2R4/4pp1r/1p2P2P/6P1/PPP2P2/1K1R4 w - - 1 24
\item 4k3/7p/5p2/4p3/Rp1r3P/5rP1/PPP1RP2/2K5 w - - 7 31
\item r2q1rk1/ppp1bp1p/2bp1np1/5NB1/4P3/2N5/PPP2PPP/R2Q1RK1 w - - 0 11
\item rn1qkb1r/1b2pppp/p2p1n2/1p6/3NP3/2NB4/PPP2PPP/R1BQ1RK1 w kq - 2 8
\item 2q2rk1/5ppp/4pn2/1B1p4/8/4QP2/2P3PP/2R3K1 w - - 0 24
\item 2r3k1/5pp1/4pn1p/1BPp4/5q2/5P1P/4Q1P1/2R3K1 w - - 1 28
\item 8/3k1p2/2p2p2/3p4/2nP1P2/p2KP1N1/PrR2P1r/2R5 w - - 2 33
\item rn1qr1k1/pp3p2/2p1b2p/3p2p1/3P2nB/2NBP1K1/PPQ1NPP1/R4R2 w - - 0 14
\item r3r1k1/pp1n1pq1/2p4p/3p4/3P1P2/3QP3/PP2N1P1/R2N1K2 w - - 1 21
\item 8/pp3pk1/2p4p/3p1P2/3Pr3/8/PP1R2P1/6K1 w - - 1 34
\item r2q1rk1/4bpp1/ppn1b2p/2pp4/3P1B2/1P2PN2/P3QPPP/1BRR2K1 w - - 2 16
\item 1b4k1/6p1/2r1qp1p/3p4/3B2P1/1P2PQ1P/5P2/3R2K1 w - - 0 34
\item 6k1/6p1/1P2qp1p/3pb3/1r1B2P1/4PQ1P/5P2/3R2K1 w - - 1 37
\item 8/8/1P4Rp/3k2p1/4pbP1/4B2P/1r3P2/5K2 w - - 2 49
\item r1b1r1k1/p1p2pb1/2pq3p/4n3/N2pPB2/3B3P/PPPQ2P1/R3R1K1 w - - 3 18
\item r1b3r1/p1p3bk/2p2p1p/2N1n2q/3pPB2/3B3P/PPPQ1RP1/5R1K w - - 4 23
\item r5r1/p1p3bk/2p4p/2N1p3/2BpP3/7q/PPPQ1R2/5R1K w - - 0 26
\item r1r5/1bp1q1k1/3p2p1/3Pp1np/1PQ3p1/4N3/1B2PRBP/5RK1 w - - 6 30
\item 8/2p2k2/3p4/3Pp1pp/1Pb1B1p1/4P1K1/1B5P/8 w - - 3 44
\end{itemize}{}

Listing two:

\begin{itemize}
\item 2r3k1/6pp/2p1q3/p2p1n2/P2Bp3/2P1P2P/6P1/R3Q1K1 w - - 0 29
\item 5rk1/1R4pp/2pq4/p2p1n2/P2Bp3/2P1P2P/6P1/5QK1 w - - 10 34
\item 8/pp2qp1k/7p/7P/1PP5/3b1BQ1/P4PK1/8 w - - 2 35
\item 5bk1/1p3p1p/2p1p1pP/p1n1P1B1/2P3P1/P2qQB2/1P3P2/6K1 w - - 1 31
\item r4rk1/1p1n2p1/1qp1pn1p/p2p2p1/1bPP2P1/4PNBP/PPQN1P2/2KR3R w - - 0 16
\item 8/1p3pkp/6p1/4p3/3r4/pP2qPP1/P1Q3KP/3R4 w - - 2 31
\item r4r2/ppp1qpkp/2bp1np1/6B1/3QP3/2N5/PPP2PPP/R4RK1 w - - 2 13
\end{itemize}{}

\section*{Appendix C}
A working implementation of this research will be made available at \url{https://github.com/ArmanMaesumi/LimitedLookAheadChess}. Our neural networks are created and trained using Keras with TensorFlow on a single NVIDIA GTX 1080Ti. The dataset was created using a multi-threaded python script, along with the Python-Chess package. Finally, the look ahead algorithm discussed in (9) was written in Golang, and will also be made available at the github repository.

\end{document}